# *DynaVINS*: A Visual-Inertial SLAM for Dynamic Environments

Seungwon Song[1], Hyungtae Lim[1], Alex Junho Lee[2] and Hyun Myung[1*], *Senior Member, IEEE*

*Abstract*—Visual inertial odometry and SLAM algorithms are widely used in various fields, such as service robots, drones, and autonomous vehicles. Most of the SLAM algorithms are based on assumption that landmarks are static. However, in the real-world, various dynamic objects exist, and they degrade the pose estimation accuracy. In addition, temporarily static objects, which are static during observation but move when they are out of sight, trigger false positive loop closings. To overcome these problems, we propose a novel visual-inertial SLAM framework, called *DynaVINS*, which is robust against both dynamic objects and temporarily static objects. In our framework, we first present a robust bundle adjustment that could reject the features from dynamic objects by leveraging pose priors estimated by the IMU preintegration. Then, a keyframe grouping and a multi-hypothesis-based constraints grouping methods are proposed to reduce the effect of temporarily static objects in the loop closing. Subsequently, we evaluated our method in a public dataset that contains numerous dynamic objects. Finally, the experimental results corroborate that our *DynaVINS* has promising performance compared with other state-of-the-art methods by successfully rejecting the effect of dynamic and temporarily static objects. Our code is available at https://github.com/url-kaist/dynaVINS.

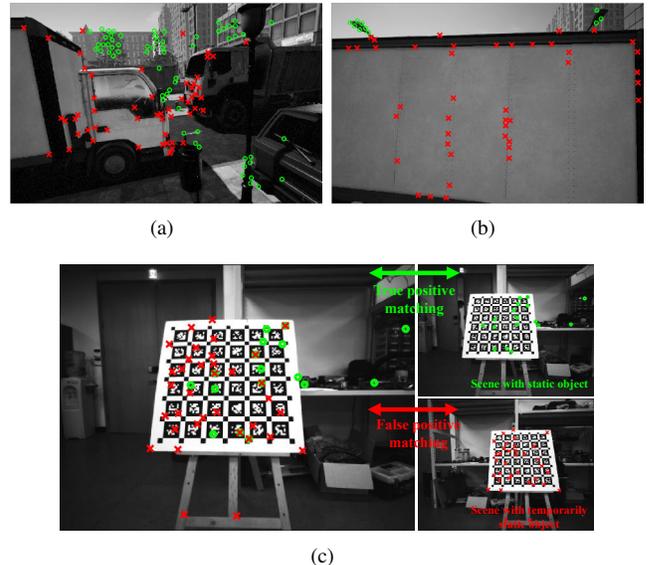

Fig. 1. Our algorithm, *DynaVINS*, in various dynamic environments. (a)–(b) Feature rejection results in `city_day` sequence of VIODE dataset [8]. Even if the most features are dynamic, *DynaVINS* can discard the effect of the dynamic features. (c) Separation of feature matching results into multiple hypotheses in `E_shape` sequence of our dataset. Even if a temporarily static object exists, only a hypothesis from static objects is determined as true positive. Features with high and low weights are denoted as green circles and red crosses, respectively, in both two cases.

## I. INTRODUCTION

Simultaneous localization and mapping (SLAM) algorithms have been widely exploited in various robotic applications that require precise positioning or navigation in environments where GPS signals are blocked. Various types of sensors have been used in SLAM algorithms [1]. In particular, visual sensors such as monocular cameras [2–4] and stereo cameras [5–7] are widely used because of their relatively low cost and weight with rich information.

Various visual SLAM methods have been studied for more than a decade. However, most researchers have assumed that landmarks are implicitly static; thus, many visual SLAM methods still have potential risks when interacting with real-world environments that contain various dynamic objects. Only recently, several studies focused on dealing with dynamic objects solely using visual sensors.

Most of the studies [9–11] address the problems by detecting the regions of dynamic objects via depth clustering, feature reprojection, or deep learning. Moreover, some researchers incorporate the dynamics of the objects into the optimization framework [12–14]. However, geometry-based methods require accurate camera poses; hence they can only deal with limited fractions of dynamic objects. In addition, deep-learning-aided methods have the limitation of solely working for predefined objects.

In the meanwhile, visual-inertial SLAM (VI-SLAM) frameworks [3–7] have been proposed by integrating an inertial measurement unit (IMU) into the visual SLAM. Unlike the visual SLAMs, a motion prior from the IMU helps the VI-SLAM algorithms to tolerate scenes with dynamic objects to some degree. However, if the dominant dynamic objects occlude most of the view as shown in Fig. 1(b), the problem cannot be solved solely using the motion prior.

In addition, in real-world applications, temporarily static objects are static while being observed but in motion when they are not under observation. These objects may lead to a critical failure in the loop closure process due to false positives as shown in Fig. 1(c). To deal with temporarily static objects, robust back-end methods [15–18] are proposed to reduce the effect of the false positive loop closures in the optimization. However, since they focused on the instantaneous false positive loop closures, they cannot deal with the persistent false positive loop closures caused by the temporarily static objects.

[1]Seungwon Song,[1]Hyungtae Lim and [1]Hyun Myung are with School of Electrical Engineering at KAIST, Daejeon, 34141, Republic of Korea.{sswan55, shapelim,hmyung}@kaist.ac.kr

[2]Alex Junho Lee is with the Department of Civil and Environmental Engineering at KAIST, Daejeon, 34141, Republic of Korea. alex_jhlee@kaist.ac.kr

\* Corresponding author: Prof. Hyun Myung

This work was supported by the "Indoor Robot Spatial AI Technology Development" project funded by KT (KT award B210000715). The students are supported by the BK21 FOUR from the Ministry of Education (Republic of Korea).

In this study, to address the aforementioned problems, we propose a robust VI-SLAM framework, called *DynaVINS*, which is robust against dynamic and temporarily static objects. Our contributions are summarized as follows:

- The robust VI-SLAM approach is proposed to handle dominant, undefined dynamic objects that cannot be solved solely by learning-based or vision-only methods.
- A novel bundle adjustment (BA) pipeline is proposed for simultaneously estimating camera poses and discarding the features from the dynamic objects that deviate significantly from the motion prior.
- A robust global optimization with constraints grouped into multiple hypotheses is proposed to reject persistent loop closures from the temporarily static objects.

In the remainder of this letter, we introduce the robust BA method for optimizing moving windows in Section III, methods for the robust global optimization in Section IV, and compare our proposed method with other state-of-the-art (SOTA) methods in various environments in Section V.

## II. RELATED WORKS

### A. Visual-inertial SLAM

As mentioned earlier, to address the limitations of the visual SLAM framework, VI-SLAM algorithms have been recently proposed to correct the scale and camera poses by adopting the IMU. MSCKF [4] was proposed as an extended Kalman filter(EKF)-based VI-SLAM algorithm. ROVIO [7] also used an EKF, but proposed a fully robocentric and direct VI-SLAM framework running in real time.

There are other approaches using optimization. OKVIS [6] proposed a keyframe-based framework and fuses the IMU preintegration residual and the reprojection residual in an optimization. ORB-SLAM3 [5] used an ORB descriptor for the feature matching, and poses and feature positions are corrected through an optimization. VINS-Fusion [3], an extended version of VINS-Mono, supports a stereo camera and adopts a feature tracking, rather than a descriptor matching, which makes the algorithm faster and more robust.

However, these VI-SLAM methods described above still have potential limitations in handling the dominant dynamic objects and the temporarily static objects.

### B. Dynamic Objects Rejection in Visual and VI SLAM

Numerous researchers have proposed various methods to handle dynamic objects in visual and VI SLAM algorithms. Fan *et al.* [10] proposed a multi-view geometry-based method using an RGB-D camera. After obtaining camera poses by minimizing the reprojection error, the type of each feature point is determined as dynamic or static by the geometric relationship between the camera movement and the feature. Canovas *et al.* [11] proposed a similar method, but adopted a surfel, similar to a polygon, to enable a real-time performance by reducing the number of items to be computed. However, multi-view geometry-based algorithms assumed that the camera pose estimation is accurate enough, leading to the failure when the camera pose estimation is inaccurate owing to the dominant dynamic objects.

One of the solutions to this problem is to employ a wheel encoder. G2P-SLAM [19] rejected loop closure matching results with a high Mahalanobis distance from the estimated pose by the wheel odometry, which is invariant to the effect of dynamic and temporarily static objects. Despite the advantages of wheel encoder, these methods are highly dependent on the wheel encoder, limiting their own applicability.

Another feasible approach is to adopt deep learning networks to identify predefined dynamic objects. In the DynaSLAM [9], masked areas of the predefined dynamic objects using a deep learning network were eliminated and the remainder was determined via multi-view geometry. In the Dynamic SLAM [20], a compensation method was adopted to make up for missed detections in a few keyframes using sequential data. Although the deep learning methods can successfully discard the dynamic objects even if they are temporarily static, these methods are somewhat problematic for the following two reasons: a) the types of dynamic objects have to be predefined, and b) sometimes, only a part of the dynamic object is visible as shown in Fig. 1(b). For these reasons, the objects may not be detected occasionally.

On the other hand, methods for tracking a dynamic object's motion have been proposed. RigidFusion [12] assumed that only a single dynamic object is in the environment and estimated the motion of the dynamic object. Qiu *et al.* [14] combined a deep learning method and VINS-Mono [3] to track poses of the camera and object simultaneously. DynaSLAM II [13] identified dynamic objects, similar to DynaSLAM [9], then, within the BA factor graph, the poses of static features and the camera were estimated while estimating the motion of the dynamic objects simultaneously.

### C. Robust Back-End

In the graph SLAM field, several researchers have attempted to discard incorrectly created constraints. For instance, max-mixture [15] employed a single integrated Bayesian framework to eliminate the incorrect loop closures, while switchable constraint [16] is proposed to adjust the weight of each constraint to eliminate false positive loop closures in the optimization. However, false-positive loop closures can be expected to be consistent and occur persistently by the temporarily static objects. These robust kernels are not appropriate to handling such persistent loop closures.

On the other hand, the Black-Rangarajan (B-R) duality [21] is proposed to unify robust estimation and outlier rejection process. Some methods [17, 18] utilize B-R duality in point cloud registration and pose graph optimization (PGO) to reduce the effect of false-positive matches even if they are dominant. These methods are useful for rejecting outliers in a PGO. However, repeatedly detected false-positive loop closures from similar objects are not considered. Moreover, B-R duality is not yet utilized in the BA of the VI-SLAM.

To address the aforementioned limitations, we improve the VI-SLAM to minimize the effect of the dynamic and temporarily static objects by adopting the B-R duality not only in the graph structure but also in the BA framework by reflecting the IMU prior and the feature tracking information.

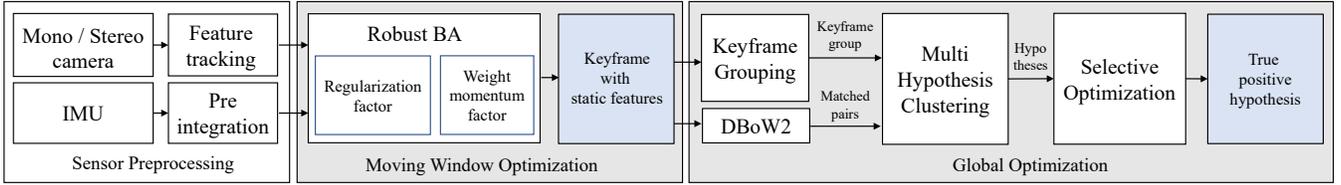

Fig. 2. The pipeline of our robust visual inertial SLAM. Features are tracked in mono or stereo images and IMU data are preintegrated in the sensor preprocessing step. Then, the robust BA is applied to discard tracked features from dynamic objects and only the features from static objects will be remain. Keyframes are grouped using the number of tracked features, and loop closures detected in current keyframe groups are clustered into hypotheses. Each hypothesis with the weight is used or rejected in the selective optimization. Using the proposed framework, a trajectory robust against dynamic and temporarily static objects can be obtained.

## III. ROBUST BUNDLE ADJUSTMENT

### A. Notation

In this letter, the following notations are defined. The $i$-th camera frame and the $j$-th tracked feature are denoted as $C_i$ and $f_j$, respectively. For two frames $C_A$ and $C_B$, $T_B^A \in SE(3)$ denotes the pose of $C_A$ relative to $C_B$. And the pose of $C_A$ in the world frame $W$ can be denoted as $T_W^A$.

$\mathcal{B}$ is a set of indices of the IMU preintegrations, and $\mathcal{P}$ is a set of visual pairs $(i, j)$ where $i$ corresponds to the frame $C_i$ and $j$ to the feature $f_j$. Because the feature $f_j$ is tracked across multiple camera frames, different camera frames can contain the same feature $f_j$. Thus, a set of indices of all tracked features in the current moving window is denoted as $\mathbf{F}_\mathcal{P}$, and a set of indices of the camera frames that contain the feature $f_j$ is denoted as $\mathcal{P}(f_j)$.

In the visual-inertial optimization framework of the current sliding window, $\mathcal{X}$ represents the full state vector that contains sets of poses and velocities of the keyframes, biases of the IMU, i.e., acceleration and gyroscope biases, and estimated depth of the features as in [3].

### B. Conventional Bundle Adjustment

In the conventional visual-inertial state estimator [3], the visual-inertial BA formulation is defined as follows:

$$\min_{\mathcal{X}} \left\{ \| \mathbf{r}_p - \mathbf{H}_p \mathcal{X} \|^2 + \sum_{k \in \mathcal{B}} \| \mathbf{r}_\mathcal{I}(\hat{\mathbf{z}}_{b_{k+1}}^{b_k}, \mathcal{X}) \|_{\mathbf{P}_{b_{k+1}}^{b_k}}^2 \right. \\ \left. + \sum_{(i,j) \in \mathcal{P}} \rho_H(\| \mathbf{r}_\mathcal{P}(\hat{\mathbf{z}}_j^{C_i}, \mathcal{X}) \|_{\mathbf{P}_j^{C_i}}^2) \right\}, \quad (1)$$

where $\rho_H(\cdot)$ denotes the Huber loss [22]; $\mathbf{r}_p$, $\mathbf{r}_\mathcal{I}$, and $\mathbf{r}_\mathcal{P}$ represent residuals for marginalization, IMU, and visual reprojection measurements, respectively; $\hat{\mathbf{z}}_{b_{k+1}}^{b_k}$ and $\hat{\mathbf{z}}_j^{C_i}$ stand for observations of IMU and feature points; $\mathbf{H}_p$ denotes a measurement estimation matrix of the marginalization, and $\mathbf{P}$ denotes the covariance of each term. For convenience, $\mathbf{r}_\mathcal{I}(\hat{\mathbf{z}}_{b_{k+1}}^{b_k}, \mathcal{X})$ and $\mathbf{r}_\mathcal{P}(\hat{\mathbf{z}}_j^{C_i}, \mathcal{X})$ are simplified as $\mathbf{r}_\mathcal{I}^k$ and $\mathbf{r}_\mathcal{P}^{j,i}$, respectively.

The Huber loss does not work successfully once the ratio of outliers increases. This is because the Huber loss does not entirely reject the residuals from outliers [23]. On the other hand, the redescending M-estimators, such as Geman-McClure (GMC) [24], ignore the outliers perfectly once the residuals are over a specific range owing to their zero-gradients. Unfortunately, this truncation triggers a problem that features considered as outliers would never become inliers even though the features are originated from static objects.

To address these problems, our BA method consists of two parts: a) a regularization factor that leverages the IMU preintegration and b) a momentum factor for considering the previous state of each weight to cover the case where the preintegration becomes temporarily inaccurate.

### C. Regularization Factor

First, to reject the outlier features while robustly estimating the poses, we propose a novel loss term inspired by the B-R duality [21] as follows:

$$\rho(w_j, \mathbf{r}_\mathcal{P}^j) = w_j^2 \mathbf{r}_\mathcal{P}^j + \lambda_w \Phi^2(w_j), \quad (2)$$

where $\mathbf{r}_\mathcal{P}^j$ denotes $\sum_{i \in \mathcal{P}(f_j)} \| \mathbf{r}_\mathcal{P}^{j,i} \|^2$ for simplicity, $w_j \in [0,1]$ denotes the weight corresponding to each feature $f_j$, and $f_j$ with $w_j$ close to 1 is determined as a static feature; $\lambda_w \in \mathbb{R}^+$ is a constant parameter; $\Phi(w_j)$ denotes the regularization factor of the weight $w_j$ and is defined as follows:

$$\Phi(w_j) = 1 - w_j. \quad (3)$$

Then, $\rho(w_j, \mathbf{r}_\mathcal{P}^j)$ in (2) is adopted instead of the Huber norm in the visual reprojection term in (1). Hence, the BA formulation can be expressed as:

$$\min_{\mathcal{X}, \mathcal{W}} \left\{ \| \mathbf{r}_p - \mathbf{H}_p \mathcal{X} \|^2 + \sum_{k \in \mathcal{B}} \| \mathbf{r}_\mathcal{I}^k \|^2 + \sum_{j \in \mathbf{F}_\mathcal{P}} \rho(w_j, \mathbf{r}_\mathcal{P}^j) \right\}, \quad (4)$$

where $\mathcal{W} = \{w_j | j \in \mathbf{F}_\mathcal{P}\}$ represents the set of all weights.

By adopting weight and regularization factor inspired by B-R duality, the influence of features with a high reprojection error compared to the estimated state can be reduced while maintaining the state estimation performance. The details will be covered in the remainder of this subsection.

(4) is solved using an alternating optimization [21]. Because the current state $\mathcal{X}$ can be estimated from the IMU preintegration and the previously optimized state, unlike other methods [17, 18], $\mathcal{W}$ is updated first with the fixed $\mathcal{X}$. Then, $\mathcal{X}$ is optimized with the fixed $\mathcal{W}$.

While optimizing $\mathcal{W}$, all terms except weights are constants. Hence, the formulation for optimizing weights can be expressed as follows:

$$\min_{\mathcal{W}} \left\{ \sum_{j \in \mathbf{F}_\mathcal{P}} \rho(w_j, \mathbf{r}_\mathcal{P}^j) \right\}. \quad (5)$$

Because the weight $w_j$ is independent to each other, (5) can be optimized independently for each $w_j$ as follows:

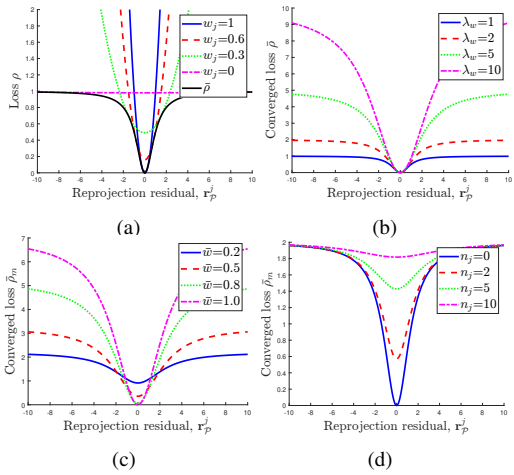

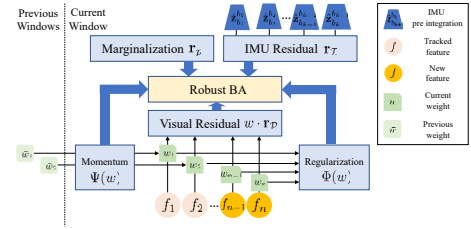

Fig. 4. Framework of robust BA. Each feature has a weight and is used in the visual residual. Each weight has been optimized through the regularization factor and the weight momentum factor. Preintegrated IMU data are used in the IMU residual term. All parameters are optimized in the robust BA.

Fig. 3. Changes of loss functions w.r.t. various parameters. (a) $\rho(w_j, \mathbf{r}_\mathcal{P}^j)$ w.r.t. $w_j$ in the alternating optimization for $\lambda_w = 1$. $\bar{\rho}(\mathbf{r}_\mathcal{P}^j)$ represents the converged loss. (b) $\bar{\rho}(\mathbf{r}_\mathcal{P}^j)$ w.r.t. $\lambda_w$. (c) $\bar{\rho}_m(\mathbf{r}_\mathcal{P}^j)$ w.r.t. $\bar{w}_j$ for $n_j = 5$. (d) $\bar{\rho}_m(\mathbf{r}_\mathcal{P}^j)$ w.r.t. $n_j$ for $\bar{w}_j = 0$.

$$\min_{w_j \in [0,1]} \left\{ w_j^2 \left( \sum_{i \in \mathcal{P}(f_j)} \|\mathbf{r}_\mathcal{P}^{j,i}\|^2 \right) + \lambda_w \Phi^2(w_j) \right\}. \quad (6)$$

Because the terms in (6) are in a quadratic form w.r.t. $w_j$, the optimal $w_j$ can be derived as follows:

$$w_j = \frac{\lambda_w}{\mathbf{r}_\mathcal{P}^j + \lambda_w}. \quad (7)$$

As mentioned previously, the weights are first optimized based on the estimated state. Thus the weights of features with high reprojection errors start with small values. However, as shown in Fig. 3(a), the loss of the feature $\rho(w_j, \mathbf{r}_\mathcal{P}^j)$ is a convex function unless the weight is zero, and there is a non-zero gradient not only in the loss of an inlier feature but also in the loss of an outlier feature, which means that the new feature affects the BA regardless of the type at first.

While the optimization step is repeated until the states and the weights are converged, the weights of the outlier features are lowered and their losses are more flattened. As a result, the losses of the outlier features approach zero-gradient and cannot affect the BA.

After convergence, the weight can be expressed using the reprojection error as in (7). Thus the converged loss $\bar{\rho}(\mathbf{r}_\mathcal{P}^j)$ can be derived by applying (7) to (2) as follows:

$$\bar{\rho}(\mathbf{r}_\mathcal{P}^j) = \frac{\lambda_w \mathbf{r}_\mathcal{P}^j}{\lambda_w + \mathbf{r}_\mathcal{P}^j}. \quad (8)$$

As shown in Fig. 3(b), increasing $\lambda_w$ affects $\bar{\rho}(\mathbf{r}_\mathcal{P}^j)$ in two directions: increasing the gradient value and convexity. By increasing the gradient value, the visual reprojection residuals affect the BA more than the marginalization and IMU preintegration residuals. And by increasing the convexity, some of the outlier features can affect the BA.

To sum up, the proposed factor benefits from both Huber loss and GMC by adjusting the weights in an adaptive way; our method efficiently filters out outliers, but does not entirely ignore outliers in the optimization at first as well.

### D. Weight Momentum Factor

When the motion becomes aggressive, the IMU preintegration becomes imprecise, and thus the estimated state becomes inaccurate. In this case, the reprojection residuals of the features from the static objects become larger; hence, by the regularization factor, those features will be ignored in the BA process even though the previous weights were close to one.

If increasing $\lambda_w$ to solve this problem, even the features with high reprojection residuals by dynamic objects are used. Therefore, the result of the BA will be inaccurate. Thus, increasing $\lambda_w$ is not enough to cope this problem.

To solve this issue, an additional factor, a weight momentum factor, is proposed to make the previously estimated feature weights unaffected by an aggressive motion.

Because the features are continuously tracked, each feature $f_j$ is optimized $n_j$ times with its previous weight $\bar{w}_j$. In order to make the current weight tend to remain at $\bar{w}_j$, and to increase the degree of the tendency as $n_j$ increases, the weight momentum factor $\Psi(w_j)$ is designed as follows:

$$\Psi(w_j) = n_j(\bar{w}_j - w_j). \quad (9)$$

Then, adding (9) to (2), the modified loss term can be derived as follows:

$$\rho_m(w_j, \mathbf{r}_\mathcal{P}^j) = w_j^2 \sum_{i \in \mathcal{P}(f_j)} \| \mathbf{r}_\mathcal{P}^{j,i} \|^2 \\ + \lambda_w \Phi^2(w_j) + \lambda_m \Psi^2(w_j), \quad (10)$$

where $\lambda_m \in \mathbb{R}^+$ represents a constant parameter to adjust the effect of the momentum factor on the BA.

In summary, proposed robust BA can be illustrated as Fig. 4. The previous weights of the tracked features are used in the weight momentum factor, and the weights of all features in the current window are used in the regularization factor. As a result, the robust BA is expressed as follows:

$$\min_{\mathcal{X}, \mathcal{W}} \left\{ \| \mathbf{r}_p - \mathbf{H}_p \mathcal{X} \|^2 + \sum_{k \in \mathcal{B}} \| \mathbf{r}_\mathcal{I}^k \|^2 + \sum_{j \in \mathbf{F}_\mathcal{P}} \rho_m(w_j, \mathbf{r}_\mathcal{P}^j) \right\}. \quad (11)$$

(11) can be solved by using the alternating optimization in the same way as (4). The alternating optimization is iterated until $\mathcal{X}$ and $\mathcal{W}$ are converged. Then, the converged loss $\bar{\rho}_m(\mathbf{r}_\mathcal{P}^j)$ can be derived. $\bar{\rho}_m(\mathbf{r}_\mathcal{P}^j)$ w.r.t. $\bar{w}_j$ and $n_j$ is shown in Fig. 3(c) and (d), respectively.

As shown in Fig. 3(c), if $\bar{w}$ is low, the gradient of the loss is small even when $\mathbf{r}_\mathcal{P}^j$ is close to 0. Thus, the features presumably originated from dynamic objects don't have much impact on the BA even if their reprojection errors are low in the current step.

Furthermore, as shown in Fig. 3(d), if $\bar{w}_j$ is zero, the gradient gets smaller as $n_j$ increases; hence the tracked outlier feature has less effect on the BA, and the longer it is tracked, the less it affects the BA.

For the stereo camera configuration, in addition to the reprojection on one camera, reprojections on the other camera in the same keyframe, $\mathbf{r}_\mathcal{P}^{\text{stereo}}$, or another keyframe, $\mathbf{r}_\mathcal{P}^{\text{another}}$, exist. In that case, weights are also applied to the reprojection $\mathbf{r}_\mathcal{P}^{\text{another}}$ because it is also affected by the movement of features, while $\mathbf{r}_\mathcal{P}^{\text{stereo}}$ is invariant to the movement of features and is only adopted as the criterion for the depth estimation.

## IV. SELECTIVE GLOBAL OPTIMIZATION

In the VIO framework, the drift is inevitably cumulative along the trajectory because the optimization is performed only within the moving window. Hence, a loop closure detection, e.g. using DBoW2 [25], is necessary to optimize all trajectories.

In a typical visual SLAM, all loop closures are exploited even if some of them are from temporarily static objects. Those false positive loop closures may lead to the failure of the SLAM framework. Moreover, features from the temporarily static objects and from the static objects may exist at the same keyframe. Therefore, in this section, we propose a method to eliminate the false positive loop closures while maintaining the true positive loop closures.

### A. Keyframe Grouping

Unlike conventional methods that treat loop closures individually, in this study, loop closures from the same features are grouped, even if they are from different keyframes. As a result, only one weight per a group is used, allowing for faster optimization.

As shown in Fig. 5(a), before grouping the loop closures, adjacent keyframes that share at least a minimum number of tracked features have to be grouped. The group starting from the $i$-th camera frame $C_i$ is defined as follows:

$$Group(C_i) = \{C_k | |\mathbf{F}_i^k| \geq \alpha, k \geq i\}, \quad (12)$$

where $\alpha$ represents a minimum number of tracked features, and $\mathbf{F}_i^k$ represents the set of features tracked from $C_i$ to $C_k$. For simplicity, $Group(C_i)$ will be denoted as $G_i$ hereinafter.

### B. Multiple Hypotheses Clustering

After keyframes are grouped as in the previous subsection, DBoW2 is employed to identify the similar keyframe $C_m$ with each keyframe $C_k$ in the current group $G_i$ starting from $C_i$ ($C_k \in G_i$ and $m < i$). Note that $C_k$ is skipped if there is no similar keyframe. After identifying up to three different $m$ for $k$, a feature matching is conducted between $C_k$ and these keyframes, and the relative pose $T_m^k$ can be obtained. Using $T_m^k$, the estimated pose of $C_k$ in the world frame, $_mT_W^k$, can be obtained as follows:

$$_mT_W^k = T_m^k \cdot T_W^m, \quad (13)$$

where $T_W^m$ represents the pose of $C_m$ in the world frame.

However, it is difficult to directly compute the similarity between the loop closures from different keyframes in the current group. Assuming that the relative pose $T_k^i$ between $C_k$ and $C_i$ is sufficiently accurate, the estimated pose of $C_i$ in the world frame can be expressed as follows:

$$_{k,m}T_W^i = T_k^i \cdot {_mT_W^k}. \quad (14)$$

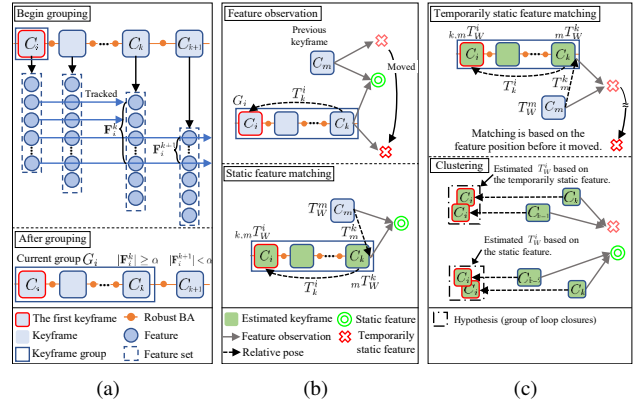

Fig. 5. The procedure of the multiple hypotheses clustering. (a) Keyframes that share the minimum number of the tracked features are grouped. (b) There are two types of features used for matchings: static and temporarily static features. $_{k,m}T_W^i$, the estimated pose of $C_i$, can be estimated using the matching result $T_m^k$ and the local relative pose $T_k^i$. An accurate keyframe pose can be estimated if static features are used for the matching. (c) The temporarily static feature is moved from the previous position. However, the matching result is based on the previous position of the feature. Thus, the estimated keyframe pose will be inaccurate. Finally, the feature matching results with similar $T_W^i$ are clustered based on the Euclidean distance.

If the features used for matchings are from the same object, the estimated $T_W^i$ of the matchings will be located close to each other, even if $C_k$ and $C_m$ of the matchings are different. Hence, after calculating Euclidean distances between the loop closure's estimated $T_W^i$, the similar loop closures with the small Euclidean distance can be clustered as shown in Fig. 5(c).

Depending on which loop closure cluster is selected, the trajectory result from the graph optimization varies. Therefore, each cluster can be called a hypothesis. To reduce the computational cost, top-two hypotheses were adopted by comparing the cardinality of the loop closures within the hypothesis. These two hypotheses of the current group $G_i$ are denoted as $H_i^0$ and $H_i^1$.

However, it is not yet possible to distinguish between true or false positive hypotheses. Hence, the method for determining the true positive hypothesis among the candidate hypotheses will be described in the next section.

### C. Selective Optimization for Constraint Groups

Most of the recent visual SLAM algorithms use a graph optimization. Let $\mathcal{C}$, $\mathcal{T}$, $\mathcal{L}$, and $\mathbf{W}$ denote the sets of keyframes, poses, loop closures, and all weights, respectively. Then the graph optimization can be denoted as:

$$\min_{\mathcal{T}} \left\{ \underbrace{\sum_{i \in \mathcal{C}} \| \mathbf{r}(T_{i+1}^i, \mathcal{T}) \|_{\mathbf{P}_{T^i}^{T^{i+1}}}^2}_{\text{local edge}} + \underbrace{\sum_{(j,k) \in \mathcal{L}} \rho_H \| \mathbf{r}(T_k^j, \mathcal{T}) \|_{\mathbf{P}_\mathcal{L}}^2}_{\text{loop closure edge}} \right\}, \quad (15)$$

where $T_{i+1}^i$ represents the local pose between two adjacent keyframes $C_i$ and $C_{i+1}$; $T_k^j$ is the relative pose between $C_j$

and $C_k$ from the loop closure; $\mathbf{P}_{T^i}^{T^{i+1}}$ and $\mathbf{P}_\mathcal{L}$ denote the covariance of the local pose and loop closure, respectively.

For the two hypotheses of group $G_i$, weights are denoted as $w_i^0$ and $w_i^1$, a sum of the weights as $\boldsymbol{w}_i$, and the set of hypotheses as $\mathcal{H}$. Using a similar procedure as in Section III.C, Black-Rangarajan duality is applied to (15) as follows:

$$\min_{\mathcal{T},\mathbf{W}} \Big\{ \sum_{i \in \mathcal{C}} \| \mathbf{r}(T_{i+1}^i, \mathcal{T}) \|^2_{\mathbf{P}_{T^i}^{T^{i+1}}}$$
$$+ \sum_{H_i \in \mathcal{H}} ((\underbrace{\sum_{(j,k) \in H_i^0} \| \frac{w_i^0}{|H_i^0|} \mathbf{r}(T_k^j, \mathcal{T}) \|^2_{\mathbf{P}_\mathcal{L}})}_{\text{residual for hypothesis 0}}$$
$$+ (\underbrace{\sum_{(j,k) \in H_i^1} \| \frac{w_i^1}{|H_i^1|} \mathbf{r}(T_k^j, \mathcal{T}) \|^2_{\mathbf{P}_\mathcal{L}})}_{\text{residual for hypothesis 1 (optional)}} + \underbrace{\lambda_l \Phi_l^2(\boldsymbol{w}_i)}_{\text{hypothesis regularization function}} )\Big\},$$
(16)

where $\lambda_l \in \mathbb{R}^+$ is a constant parameter. The regularization factor for the loop closure, $\Phi_l$, is defined as follows:

$$\Phi_l(\boldsymbol{w}_i) = 1 - \boldsymbol{w}_i = 1 - (w_i^0 + w_i^1), \quad (17)$$

where $w_i^0, w_i^1 \in [0, 1]$. To ensure that the weights are not affected by the number of loop closures in the hypothesis, the weights are divided by the cardinality of each hypothesis.

Then, (16) is optimized in the same manner as (11). Accordingly, only the hypothesis with a high weight is adopted in the optimization. In addition, all weights can be close to 0 when all hypotheses are false positives due to the multiple temporarily static objects. Hence, the failure caused by false positive hypotheses can be prevented.

Because keyframe poses are changed after the optimization, the hypothesis clustering in Section IV.B is conducted again for all groups for the next optimization.

## V. EXPERIMENTAL RESULTS

To evaluate the proposed algorithm, we compare ours with SOTA algorithms, namely, *VINS-Fusion* [3], *ORB-SLAM3* [5], and *DynaSLAM* [9]. Each algorithm is tested in a mono-inertial (`-M-I`) and a stereo-inertial (`-S-I`) mode. Note that an IMU is not used in DynaSLAM, so it is only tested in a stereo (`-S`) mode and compared with the `-S-I` mode of other algorithms. It could be somewhat unfair, but the comparison is conducted to stress the necessity for an IMU when dealing with dynamic environments.

### A. Dataset

**VIODE Dataset** VIODE dataset [8] is a simulated dataset that contains lots of moving objects, such as cars or trucks, compared with conventional datasets. In addition, the dataset includes overall occlusion situations, where most parts of the image are occluded by dominant dynamic objects as shown in Fig. 1. Note that the sub-sequence name `none` to `high` means how many dynamic objects exist in the scene.

**Our Dataset** Unfortunately, VIODE dataset does not contain harsh loop closing situations caused by temporarily static objects. Accordingly, we obtained our dataset with four sequences to evaluate our global optimization. First, `Static` sequence validates the dataset. In `Dynamic follow` sequence, a dominant dynamic object moves in front of the camera. Next, in `Temporal static` sequence, the same object is observed from multiple locations. In other words, the object is static while being observed, and then it moves to a different position. Finally, in `E-shape` sequence, the camera moves along the shape of the letter E. The checkerboard is moved while not being observed, thus it will be observed at the three end-vertices of the E-shaped trajectory in the camera perspective, which triggers the false-positive loop closures. Note that the feature-rich checkerboard is used in the experiment to address the effect of false loop closures.

### B. Error Metrics

The accuracy of the estimated trajectory from each algorithm is measured by Absolute Trajectory Error (ATE) [26], which directly measures the difference between points of the ground truth and the aligned estimated trajectory. In addition, for the VIODE dataset, the degradation rate [8], $r_d = \text{ATE}_{\text{high}}/\text{ATE}_{\text{none}}$, is calculated to determine the robustness of the algorithm.

### C. Evaluation on the VIODE Dataset

First, the effects of the proposed factors on the BA time cost and accuracy are analyzed as shown in Table I. Ours with only the regularization factor has a better result than VINS-Fusion, but with the momentum factor together, it shows not only the outperforming result than VINS-Fusion, but also the less time due to a previous information. Moreover, although the BA time of ours was increased due to additional optimization steps, it is sufficient for high-level control of robots.

As shown in Table II and Fig. 6, the SOTA methods show precise pose estimation results in static environments. However, they struggle with the effect of dominant dynamic objects. In particular, even though DynaSLAM employs a semantic segmentation module, DynaSLAM tends to diverge or shows large ATE compared with other methods as the number of dynamic objects increases (from `none` to `high`). This performance degradation is due to the overall occlusion situations, leading to the failure of the semantic segmentation module and the absence of features from static objects.

Similarly, although ORB-SLAM3 tries to reject the frames with inaccurate features, it diverges when dominant dynamic objects exist in `parking_lot mid`, `high` and `city_day high` sequences. However, especially in `parking_lot low` sequence, there is only one vehicle that is far from the camera, and it occludes an unnecessary background environment. As a consequence, ORB-SLAM3-`S-I` outperforms other algorithms.

VINS-Fusion is less hindered by the dynamic objects because it tries to remove the features with an incorrectly

TABLE I. Ablation experiment tested in `parking_lot high` sequence in VIODE dataset [8]. All algorithms are set to `-S-I` mode.

| Method | ATE (m) | Average BA time (ms) |
|---|---|---|
| VINS-Fusion | 0.2780 | 30.9134 |
| DynaVINS (Regularization) | 0.0972 | 60.7384 |
| DynaVINS (Regularization + Momentum) | 0.0416 | 53.0432 |

TABLE II. Comparison with state-of-the-art methods (RMSE of ATE in [m]). *: Failure case (diverged), `-M-I`: Mono-inertial mode, `-S`: Stereo mode, `-S-I`: Stereo-inertial mode, `SC`: Switchable Constraints [16] Parameters for DynaVINS in VIODE: $\lambda_w = 1.0, \lambda_m = 0.2$ and in our dataset: $\lambda_w = 1.0, \lambda_m = 1.0, \lambda_l = 1.0$.

| Method | VIODE [8] | | | | | | | | | | | | Our dataset | | | |
|---|---|---|---|---|---|---|---|---|---|---|---|---|---|---|---|---|
| | city_day | | | | city_night | | | | parking_lot | | | | Static | Dynamic follow | Temporal static | E-shape |
| | none | low | mid | high | none | low | mid | high | none | low | mid | high | | | | |
| ORB-SLAM3-`M-I` | 1.940 | 0.857 | 4.486 | * | * | * | * | * | 0.147 | 0.175 | 0.145 | 0.194 | 0.379 | 1.374 | 0.775 | * |
| VINS-Fusion-`M-I` | **0.210** | 0.182 | 0.560 | 0.510 | 0.328 | 0.371 | 0.457 | 0.464 | 0.102 | 0.138 | 0.707 | 1.135 | 0.080 | 0.463 | 0.414 | 0.727 |
| VINS-Fusion-`M-I` with `SC` | | | | | | | | | | | | | | | 0.091 | 0.736 |
| **DynaVINS-`M-I`** | 0.224 | **0.167** | **0.154** | **0.364** | **0.189** | **0.181** | **0.184** | **0.256** | **0.097** | **0.120** | **0.118** | **0.149** | **0.048** | **0.141** | **0.051** | **0.107** |
| DynaSLAM-`S` | 1.621 | 1.426 | 1.638 | * | 3.333 | 3.314 | 3.074 | 3.865 | 0.108 | 0.170 | * | * | 0.081 | 1.017 | 0.467 | 0.937 |
| ORB-SLAM3-`S-I` | 0.302 | 0.419 | 0.217 | * | 0.709 | 0.895 | 1.693 | 3.006 | 0.148 | 0.067 | * | * | 0.069 | * | 0.067 | 0.476 |
| VINS-Fusion-`S-I` | **0.150** | 0.203 | 0.234 | 0.373 | 0.317 | 0.507 | 0.494 | 0.828 | 0.121 | 0.121 | 0.212 | 0.278 | **0.029** | 0.383 | 0.229 | 0.711 |
| VINS-Fusion-`S-I` with `SC` | | | | | | | | | | | | | | | 0.034 | 0.686 |
| **DynaVINS-`S-I`** | 0.171 | **0.178** | **0.091** | **0.148** | **0.213** | **0.182** | **0.201** | **0.198** | **0.049** | **0.042** | **0.064** | **0.042** | 0.032 | **0.038** | **0.025** | **0.029** |

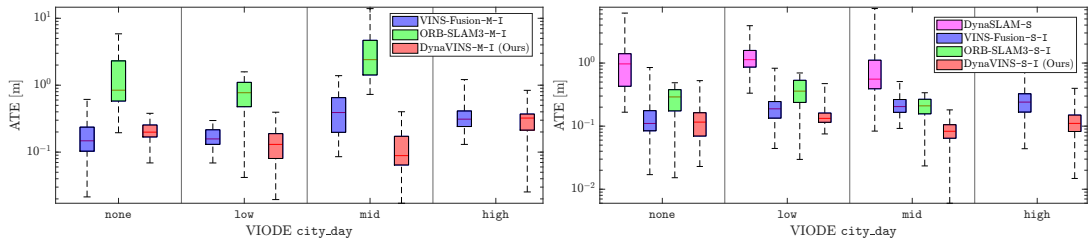

(a) Mono-inertial mode  (b) Stereo and stereo-inertial mode

Fig. 6. ATE results of state-of-the-art algorithms and ours on the `city_day` sequences of the VIODE dataset [8]. Note that the *y*-axis is expressed in logarithmic scale. Our algorithm shows promising performance with less performance degeneration compared with the other state-of-the-art methods.

TABLE III. Comparison of degradation rate $r_d$.

| Method | VIODE [8] | | |
|---|---|---|---|
| | city_day | city_night | parking_lot |
| VINS-Fusion-`M-I` | 2.425 | 1.412 | 11.167 |
| ORB-SLAM3-`M-I` | * | * | 1.693 |
| DynaSLAM-`S` | * | **1.160** | * |
| **DynaVINS-`M-I`** | **1.625** | 1.360 | **1.531** |
| VINS-Fusion-`S-I` | 2.485 | 2.613 | 1.511 |
| ORB-SLAM3-`S-I` | * | 4.238 | * |
| **DynaVINS-`S-I`** | **0.864** | **0.929** | **0.857** |

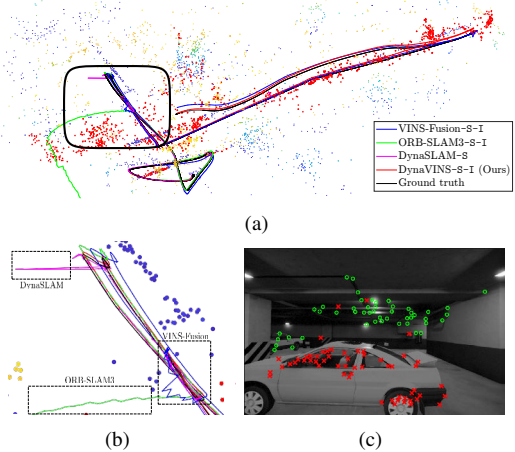

Fig. 7. Results of the state-of-the-art algorithms and ours on the `parking_lot high` sequence of the VIODE dataset [8]. (a) Trajectory of each algorithm in the 3D feature map, which is the result of our proposed algorithm. Features with low weight are depicted in red. (b) Enlarged view of (a). All other algorithms except our algorithm lost track or had noisy trajectories while observing dynamic objects and as in (c) feature weighting result of our algorithm, features from dynamic objects (red crosses) have low weight while robust features (green circles) have high weight.

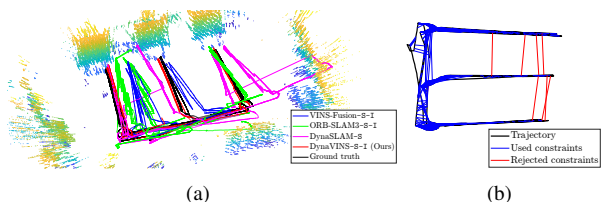

Fig. 8. Results of the algorithms on `E-shape` sequence. (a) Trajectory results. Other algorithms are inaccurate due to false positive loop closures. (b) A loop closure rejection result of our algorithm. Constraints with low weight (red lines) do not contribute to the optimized trajectory.

estimated depth (negative or far) after BA. However, those features have affected the BA before they are removed. As a result, as the number of the features from dynamic objects increases, the trajectory error of VINS-Fusion gets higher.

In contrast, our proposed method shows promising performance in both mono-inertial and stereo-inertial modes. For example, in `parking_lot high` sequence as shown in Fig. 7(a)–(b), ours performs stable pose estimation even when other algorithms are influenced by dynamic objects. Moreover, even though the number of dynamic objects increases, a performance degradation remains small compared to other methods in all scenes. This confirms that our method overcomes the problems caused by dynamic objects owing to our robust BA method, which is also supported by Table III. In other words, our proposed method successfully rejects all dynamic features by adjusting the weights in an adaptive way. Also, our method could be even robust against the overall occlusion situations, as shown in Fig. 1(b).

Interestingly, our proposed robust BA method enables robustness against changes in illuminance by rejecting the inconsistent features (e.g., low weight features in dark area of Fig. 7(c)). Accordingly, our method shows remarkable performance compared with the SOTA methods in `city_night` scenes where not only dynamic objects exist, but also there is a lack of illuminance. Note that `-M-I` of ours has better result than `-S-I`. This is because the stereo reprojection, $\mathbf{r}_{\mathcal{P}}^{\text{stereo}}$, can be inaccurate in low-light conditions.

*D. Evaluation on Our Dataset*

In the `static` case, all algorithms have low ATE values. This sequence validates that our dataset is correctly obtained.

However, in `Dynamic follow`, other algorithms tried to track the occluding object. Hence, not only failures of BA but also false-positive loop closures are triggered. Consequently, other algorithms except ours have higher ATEs.

Furthermore, in `Temporal static`, ORB-SLAM3 and VINS-Fusion can eliminate the false-positive loop closure in the stereo-inertial case. However, in the mono-inertial case, due to an inaccurate depth estimation, they cannot reject the false-positive loop closures. Additionaly, VINS-Fusion with Switchable Constraints [16] can also reject the false-positive loop closures, but ours has a better performance as shown in Table II.

Finally, in `E-shape` case, other algorithms fail to optimize the trajectory, as illustrated in Fig. 8(a), owing to the false-positive loop closures. Also VINS-Fusion with Switchable Constraints cannot reject the false-positive loop closures that are continuously generated. However, ours optimizes the weight of each hypothesis, not individual loop closures. Hence, false-positive loop closures are rejected in the optimization irrespective of the number of them, as illustrated in Fig. 8(b). Ours does not use any object-wise information from the image; hence the features from the same object can be divided into different hypotheses, as depicted in Fig. 1(c).

## VI. CONCLUSIONS

In this study, DynaVINS has been proposed, which is a robust visual-inertial SLAM framework based on the robust BA and the selective global optimization in dynamic environments. The experimental evidence corroborated that our algorithm works better than other algorithms in simulations and in actual environments with various dynamic objects. In future works, we plan to improve the speed and the performance. Moreover, we will adopt the concept of DynaVINS to the LiDAR-Visual-Inertial (LVI) SLAM framework.